\documentclass{amsart}

\usepackage{amsmath}
\usepackage{amsthm}
\usepackage{amsfonts}
\usepackage{amssymb}
\usepackage{epsfig}
\usepackage{graphicx} 
\usepackage{palatino,url}

\theoremstyle{plain}

\theoremstyle{remark}

\theoremstyle{definition}

\include{psfig}

\DeclareMathOperator*{\argmax}{argmax}

\begin{document}
\title{Fully Automatic 3D Reconstruction of Histological Images}

\author[Ba\u{g}c\i]{Ula\c{s} Ba\u{g}c\i}
\address{Collaborative Medical Image Analysis on Grid (CMIAG),The University of Nottingham, Nottingham, UK}
\email{ulasbagci@ieee.org}
\urladdr{www.ulasbagci.net}

\author[Bai]{Li Bai}

\begin{abstract}
In this paper, we propose a computational framework for 3D volume reconstruction from 2D histological slices using  registration algorithms in feature space. To improve the quality of reconstructed 3D volume, first, intensity variations in images are corrected by an intensity standardization process which maps image intensity scale to a standard scale where similar intensities correspond to similar tissues. Second, a subvolume approach is proposed for 3D reconstruction by dividing standardized slices into groups.
Third,  in order to improve the quality of the reconstruction process, an automatic best reference slice selection algorithm is developed based on an iterative assessment of image entropy and mean square error of the registration process. Finally, we demonstrate that the choice of the reference slice has a significant impact on registration quality and subsequent 3D reconstruction.
\end{abstract}
\maketitle 
\section{Introduction}
\label{sec:intro}
2D imaging methods, such as optical microscopy, are still preferable to 3D imaging methods due to their high level of specificity and high resolution properties. Histological sections (slices) obtained through 2D imaging methods provide useful information for the diagnosis or the study of pathology. Although 2D histological slices have great impacts on quantification and visualization of clinical data, 3D volume reconstruction from these 2D slices is required in order to fully appreciate anatomical structures~\cite{ourselin00}. 

Typically, a 3D volume is reconstructed by registering (aligning) the 2D sections with respect to a chosen reference and stacking successive aligned sections~\cite{malandain04}. As the acquisition processes of different 2D histological images are performed independently, slice misalignment and deformation is often unavoidable. The deformation varies from section to section and non-cohorent distortions may exist in consecutive sections. Choosing an arbitrary slice as a reference slice leads to errors in 3D volume reconstruction, hence,  the reference slice should be chosen properly not to contain distortions in order to achive high quality volume reconstruction~\cite{bagci_spie08, bagci_report}.

Fully automatic registration of histological slices and reconstruction of 3D volume are necessary for two reasons. First, since manual registration using interactive alignment is non-reproducible and user dependent, it cannot be used if the number of slices is large~\cite{ourselin00,malandain04}. To quantify changes between images, motion and deformation characteristics specify the type of transformation (registration). Since histological slices change smoothly from slice to slice and the section distortions induced by the preparation process are local in nature~\cite{bagci_spie08, ju06, bagci_report}, accurate alignment of these slices can be achived by using elastic registration methods. Second, since manual selection of the best reference slice (BRS) uses qualitative measures and ignores the image information content, optimum smooth 3D volume reconstruction cannot be guaranteed.

In this paper, we present a fully automatic 3D reconstruction method which  tackles three difficult problems in registration of histological images. Section~\ref{sec:std} explains an important preprocessing method, called \textit{standardization}, which captures intensity variations between slices and plays a significant role in identifying BRS selection. In Section~\ref{sec:feat}, the  \textit{edgeness} space is presented for the registration framework to provide better global alignment and to avoid possible misalignments. Section~\ref{sec:affine} briefly explains locally affine globally smooth (LAGS) registration method. Based on iterative assessment of image entropy and Mean Square Error (MSE) of the registration process in feature space, an automatic BRS selection algorithm is described in Section~\ref{sec:bestref}. To evaluate reconstructed volume qualitatively and quantitatively, we use both Correlation Alignment Measure (CAM) and proposed Standard Deviation Maps (SDM) in Section~\ref{sec:results}. Evaluations and discussions are given in Section~\ref{sec:conc}. 

\section{Standardization of Image Intensity Scale}
\label{sec:std}
Image intensity variations are not only influenced by the distribution of light sources, but also the content (different tissues) of the images as different tissues show different intensity levels. To avoid the effects of illumination conditions and identify those intensity variations due to different tissue types, a standardization procedure is applied to histological images. 

Standardization is a non-linear pre-processing technique which maps image intensity histogram (scale) into a standard intensity histogram (scale) so that similar intensities will have similar tissue meaning after standardization. Standardization was firstly developed for MR images~\cite{udupa_std_jmri}. In previous works~\cite{bagci_spie08, bagci_report}, we applied this method to standardize histological images. 

%Figure~\ref{img:beforeandafter} shows slices before and after standardization. The first row shows the original data displayed using the default window setting. The second row shows the same slices after standardization displayed using the "standard" window settings with the parameters defined in~\cite{bagci_spie08, udupa_std_jmri, bagci07, bagci_report}.

%\begin{figure}[h]
%\begin{minipage}[b]{1.0\linewidth}
% \centerline{\epsfig{figure=beforeandafterstandardization.eps,width=9.5cm}}
%\caption{Various Original slices before (first row) and after (second row) standardization\label{img:beforeandafter}}
%\end{minipage}
%\end{figure} 

\section{Feature Space}
\label{sec:feat}
Choice of the feature space plays a significant role in image registration especially if the similarity metric is based on the optimization function independent of spatial information such as mutual information. Since these kind of registration methods do not take into consideration the spatial information of pixel/voxel intensity distribution/variation, the optimization algorithm may get stuck in local maximum resulting in misalignment. Defining a feature space capturing variations of gray-level characteristics will overcome the drawbacks of intensity based approaches. To align the images globally, we used a particular feature space which represents an image by continuous variables, called edgeness, and describes the intensity variance of a predefined region over the image~\cite{thiran01, bagci_spie08, bagci_report}.

We represent an image (section/slice/scene) by a pair $\mathbf{F}=(F,g)$ where $F$ is a two-dimensional (2-D) array of scene elements (pixels) and  $g$ is intensity function, whose domain is $F$. We assign an integer intensity value for each pixel $o \in \mathbf{F}$. Edgeness feature space is defined by the pair $\mathbf{F_e}=(\mathbf{F},r_f)=(F,g,r_f)$, where $r_f$ is a fixed radius for each region. At image coordinate $r_0$, the edgeness is represented by
\begin{equation}
\label{eq:edgeness}
\mathbf{F_e} = \sum_{|r_i - r_0|<r_f} | g(r_i) - g(r_0)|,
\end{equation} 
where $r_f$ is the radius. It should not be concluded that the process is just emphasizing edges and deciding whether a specific voxel/pixel belongs to the edge or not~\cite{thiran01, bagci_spie08, bagci_report}. Instead, within a specified radius value, the image feature content is forced to stay beyond a variation level which prevents the registration process from getting stuck in local maxima.

\section{Locally Affine Nonlinear Transformation}
\label{sec:affine}
Local alignment (elastic) of images is obtained through Locally Affine Globall Smooth (LAGS)
registration method described in~\cite{periaswamy,bagci07, bagci_spie08}. Since consecutive slices are not exactly the same, rather slices vary smoothly, LAGS registration algorithm fits well to the problem. For 2D images, 8 affine parameters are needed to fully identify changes between images. Two of these affine parameters are needed to capture local brightness and contrast patterns~\cite{periaswamy} and 6 affine parameters are used to capture local deformations for 2D images. Since the standardization procedure has been used to remove intensity variations among the same tissue types, there is no need to use these 2 affine parameters. Briefly, LAGS registration algorithm uses the difference image of source and target image as an optimization function and tries to minimize it over small local image domain. Readers are strongly encouraged to read~\cite{periaswamy,bagci07, bagci_spie08, bagci_report} to understand the theory of LAGS and the modified algorithm which takes into account the standardization procedure. 

\section{Automatic Best Reference Slice Selection}
\label{sec:bestref}
The quality of the 3D volume reconstruction process mostly depends on the choice of the reference slice. The reference slice is used as a target image and all the remaining slices are being considered as source images to be registered onto the target image. If the reference slice is distorted or noisy, reconstructed 3D volume will not be optimal. Once the reference slice is identified as target image, registration based fusion methodology can be applied for reconstruction~\cite{malandain04}.  

Selecting best reference slice can be based on high confidence image features such as MSE, entropy, edge, texture, color, intensity histograms, etc.
\begin{enumerate}
\item \textbf{MSE:} In the case of distortions, structural discontinuity is not minimum even for the consecutive slices. When affine registration is performed for global alignment of images, the optimization procedure tries to minimize MSE between images but due to distortion, it will not reach low MSE values. Furthermore, it is also known that with small SNR values, alignment is difficult, leading to high registration errors. Therefore, MSE can be used as a tool for checking whether the slices are distorted or not. While high MSE values indicate most probably distorted and noisy slices, low MSE values indicate strong similarity between consecutive images. 

\item \textbf{Edge:} In feature space, we emphasise edgeness
features of an image by mapping image space into the feature space where edgeness parameters hold both edge information and spatial variations of pixel intensities over all regions in the image. Therefore, we assume that MSE between any image pair already includes high confidence information related to edges.

\item \textbf{Contrast/Brightness:} Contrast/brightness patterns also play an important role in image contents. Since the standardization method has been used to correct intensity variations, intensity for the same tissue is the same for all images. 
 
\item \textbf{Entropy:} Entropy is another measure often used to characterise the information content of a data source. It has been used as a metric for image registration in the form of mutual information. Large mutual information between images implies high similarity and vice versa.
 % While BRS has been chosen, there is a strong possibility that end of slices may contain little information and feature space based registration error MSE may become small even those slices are not suitable for reference slice for 3D reconstruction. To avoid such mis-selections, entropy information can be used for high confidence feature together with MSE.
\end{enumerate}

To select BRS automatically, one needs to define a metric which describes the reference slice in terms of noise, distortion and information content levels. As in~\cite{bagci_spie08, bagci_report}, BRS can be formulated as
\begin{equation}
BRS = \argmax_{i\neq j\in V}  \left\lbrace log\left( \frac{E^{(j)}}{MSE_{i,j}}\right) \right\rbrace, 
\end{equation} 
where $MSE_{i,j}$ is the feature space based mean square error after registering the image $i$ into the image $j$ and $E$ is entropy. Further theoretical details of BRS selection algorithm can be found in~\cite{bagci_spie08, bagci_report}.

\section{Implementation, Experiments and Results}
\label{sec:results}
Registration of histological slices requires serial registration procedure which is just a combination of transformation functions.
Let $\left\lbrace A_{j \leftarrow i}| i<j\right\rbrace $ be the transformation function that warps the source image $i$ into the target image $j$. The transformation $A_{j \leftarrow i}$ is computed serially as follows
\begin{eqnarray}
A_{j \leftarrow i} = A_{j \leftarrow j-1} \circ  A_{j-1 \leftarrow j-2} \circ \ldots A_{i+1 \leftarrow i},\qquad for  \quad i<j\nonumber \\
A_{j \leftarrow i} = A_{j \leftarrow j+1} \circ  A_{j+1 \leftarrow j+2} \circ \ldots A_{i-1 \leftarrow i},\qquad for \quad  i>j,
\end{eqnarray}
where $\circ$ represents the composition.

 One of the advantages of using the subvolume approach is to avoid  the "banana shape" effect resulting from the reconstuction process~\cite{ourselin00,malandain04}. In serial registration, one of the disadvantages of using affine registration is that we lose the topology of the reconstructed volume. To reduce or eliminate this effect, one may need to use MRI of the volume superimposed onto the reconstructed histological volume or use rigid registration. 
  
In summary, registration is performed initially for slices in each subvolume separately. Three kinds of registration are performed in the reconstruction process: rigid, affine and LAGS. MSEs are calculated according to affine registration in edgeness space and have been used to select BRSs for each subvolume. Affine registration is performed in a serial manner combining transformation functions.  Then, LAGS registration is performed to capture local deformations in each subvolume with respect to the chosen reference. Once LAGS registration has been finished, subvolumes are registered to each other in a rigid manner. 

\subsection{Evaluations and Results}
We have registered a stack of 350 Nissl-stained slices acquired by cyro-sectioning coronally on an adult mouse brain with a resolution of 590x520 pixels at a resolution of 15$\mu$m and 24-bit color format \cite{ratatlas}. 

Quantitative evaluation of the results of the reconstruction process is often difficult. It has been shown in~\cite{guest01} that an ideal measure of the quality of the reconstruction is the smoothness of the reconstructed surfaces. In this work, they propose a new measure based on evaluation of smoothness of the reconstructed volume called \textit{Correspondence Alignment Measure} (CAM). As an alternative method to CAM,  \textit{Standard Deviation Maps} (SDM) is proposed to measure the smoothness of the reconstructed volume. The method is based on the standard deviation of the pixel values for the same location in each section. The CAM and SDM results for the reconstructed 3D volume and validation of the SDM are given in the following subsections.  
\subsubsection{CAM}
The CAM measure relies on the assumption that if a point is perfectly aligned, it lies midway between its corresponding points on neighbors' sections. To compute the CAM measure for a given image, first of all, corresponding points for specifed control points in the image are identified. The associated confidence values in two adjacent images are then calculated. If the confidence is  greater than a pre-defined threshold $\tau$, square root of the summation of the deformation vectors are added to the cumulative sum. Finally, the cumulative sum is normalized by the number of pixels which have contributed.  Note that CAM gives one value for each image, therefore, mean or standard deviation of CAM values of serial images are needed to compare reconstructions. Reconstucted volume is smooth if the mean or the standard deviation of CAM measures are low and vice versa. 

Summary of the changes in mean and standard deviation in CAM values is given by Table~\ref{table:cam}. The values in Table~\ref{table:cam} are obtained by considering the worst case which uses all the slices instead of just a few slices from the middle of the stack as defined in~\cite{guest01}, and $\tau$ is set to $0$. Even for the worst case, CAM values indicate that a smooth volume is constructed with the proposed framework. While mean values dropped by $7.29$\% and $18.69$\%, the standard deviation values dropped by $24.46$\% and $27.73$\% for affine and locally affine registered stacks respectively.

\begin{figure}[h]
\includegraphics[scale=0.6]{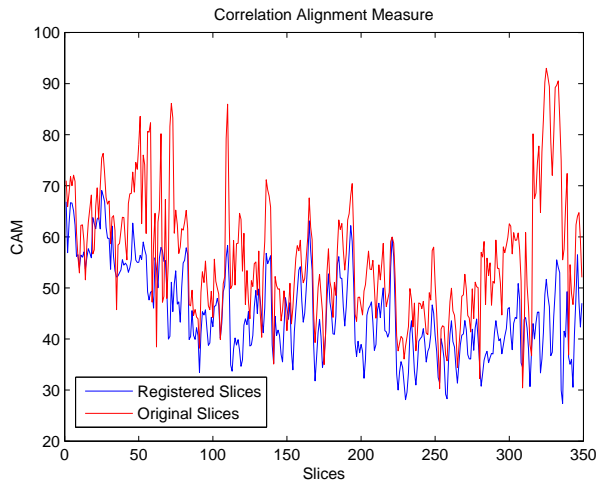} 
\caption{CAM for original and registered slices \label{img:cam}}
\end{figure}

\begin{table}[h]
\begin{center}
\caption{CAM-mean and standard deviation values for reconstructed 3D\label{table:cam}}
\begin{tabular}{|c|c|c|c|}
\hline
-- & Rigid Reg.& Affine Reg. & LAGS Reg. \\ \hline 
Mean                    &  55.911         & 51.832  & 45.461 \\  \hline
Std                       &  12.223         &  9.232    & 8.833  \\ \hline
\end{tabular}
\end{center}
\end{table}

\subsubsection{SDM}
We offer here a simple way of measuring the quality of the reconstruction by considering the smoothness of the reconstructed surfaces. If the reconstructed volume is naturally smooth, it means that the structures change smoothly and slowly from slice to slice which highly depends on the registration quality.

\textbf{Assumption:} For any reconstucted volume $\mathcal{V}$, which is a sequence of images $\mathbf{F^{(j)}}$, $j=1,..,M$, if we take the same pixel position $v$ for all slices in $\mathbf{F}$, one should expect smooth transition of pixel values within the slices of $\mathbf{F}$, if the slices are well registered. 

To validate this assumption experimentally, a 3D MRI volume $\mathcal{MV}$ is taken and its slices are warped  by applying  random deformations spanning high-to-low level. The randomly warped slices are used to reconstruct a warped 3D MRI volume $\mathcal{MV}^w$  for which we compute SDM and compare it with the smoothness of $\mathcal{MV}$.

\begin{figure}[h]
\begin{minipage}[b]{1\linewidth}
  \centering
 \centerline{\epsfig{figure=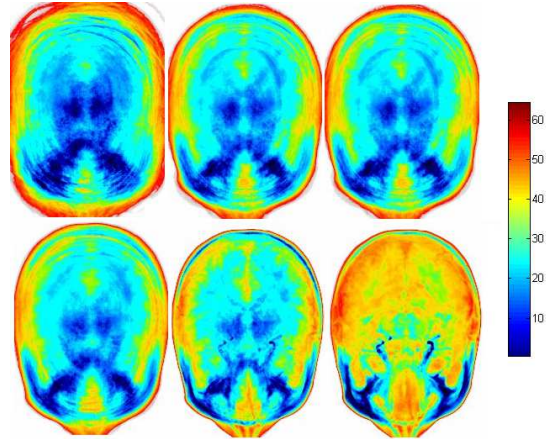,width=9.0cm}}
\caption{SDMs for warped volumes spanning from high-to-low level of deformations \label{img:testsmooth}}
\end{minipage}
\end{figure}

\begin{figure}[h]
\begin{minipage}[b]{1\linewidth}
  \centering
 \centerline{\epsfig{figure=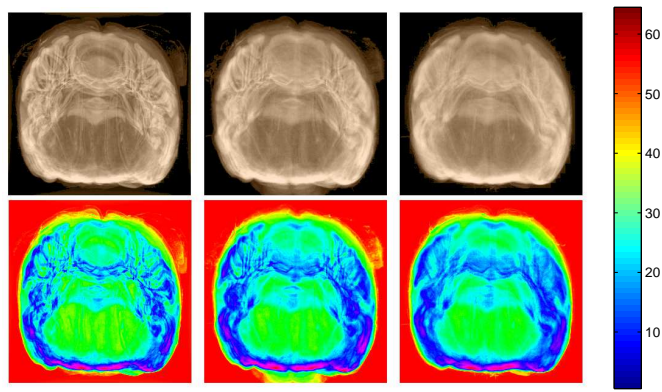,width=11.0cm}}
\caption{Standard Deviation Maps for Rigid registered, Affine-Registered and Elastic-Registered Stacks \label{img:smooth}}
\end{minipage}
\end{figure}

Figure~\ref{img:testsmooth} shows the experimental validation of SDM for different level of deformations applied to the slices of $\mathcal{MV}$. While the first SDM (the first image in the first row) has carried the highest deformation level, the level of deformation has been decreased until the last SDM (the third image in the second row) is obtained, which carries no deformation level, $\mathcal{MV}$ itself. As the deformation level decreases, the smoothness level of $\mathcal{MV}^w$ approaches the original volume $\mathcal{MV}$.
SDMs for the reconstructed volume $\mathcal{MV}^w$ for $w=$rigid, affine and LAGS are shown in Figure~\ref{img:smooth}, both in gray scale and spectrum format. Although the volume reconstructed by successive affine registrations is smoother than the volume reconstructed by successive rigid registrations, it includes the "banana-shape" effect which can be corrected by superimposing MRI of the rat brain (if available) on the reconstructed volume. Among three methods, the smoothest reconstructed volume is obtained by the proposed method shown in the last SDM in Figure~\ref{img:smooth}. 

\section{Conclusion}
\label{sec:conc}
In this paper, we have presented a novel framework to reconstruct  3D rat brain volume from 2D histological images. The framework is based on three fundamental premises. (1) All histological images must be standardized for accurate registration leading to 3D volume reconstruction. (2) For accurate and succesful registrations in consecutive slices, a reliable feature space must be taken into account. (3) For automatic 3D volume reconstruction, the reference slice must be chosen properly by avoiding slices with high noise, distortions and other factors. To validate the reconstructed volume, the smoothness of the volume is considered. In addition to the existing method CAM, we have proposed  a method called SDM to measure the smoothness of the reconstructed volume. Qualitative and quantitave evaluation of experimental results indicate that the reconstructed volume is highly accurate.

%\begin{figure}[htb]
%\begin{minipage}[b]{1.0\linewidth}
%  \centering
% \centerline{\epsfig{figure=image1.ps,width=8.5cm}}
%  \vspace{2.0cm}
%  \centerline{(a) Result 1}\medskip
%\end{minipage}
%
%\begin{minipage}[b]{.48\linewidth}
%  \centering
% \centerline{\epsfig{figure=image3.ps,width=4.0cm}}
%  \vspace{1.5cm}
%  \centerline{(b) Results 3}\medskip
%\end{minipage}
%\hfill
%\begin{minipage}[b]{0.48\linewidth}
%  \centering
% \centerline{\epsfig{figure=image4.ps,width=4.0cm}}
%  \vspace{1.5cm}
%  \centerline{(c) Result 4}\medskip
%\end{minipage}
%
%\caption{Example of placing a figure with experimental results.}
%\label{fig:res}
%
%\end{figure}

%\vfill
%\pagebreak

\section{Acknowledgements}
We wish to thank Prof. Jayaram K. Udupa, from Medical Image Processing Group of University of Pennsylvania, for his encouragement and guidance throughout this project.

%\bibliographystyle{IEEEbib}
%\bibliography{refs}

\end{document}